\documentclass[10pt,twocolumn,letterpaper]{article}

\usepackage{wacv}
\usepackage{times}
\usepackage{epsfig}
\usepackage{graphicx}
\usepackage{amsmath}
\usepackage{amssymb}
\usepackage{comment}
\usepackage{booktabs}
\usepackage{threeparttable}
\usepackage{multirow}
\usepackage{diagbox}
\usepackage{enumitem}
\usepackage{cite}

\def\wacvPaperID{1855} 

\wacvapplicationstrack   

\wacvfinalcopy 

\ifwacvfinal
\usepackage[breaklinks=true,bookmarks=false]{hyperref}
\usepackage{breakcites}
\else
\usepackage[pagebackref=true,breaklinks=true,colorlinks,bookmarks=false]{hyperref}
\usepackage{breakcites}
\fi

\begin{document}

\title{Enabling ISP-less Low-Power Computer Vision}

\author{Gourav Datta, Zeyu Liu, Zihan Yin, Linyu Sun, Akhilesh R. Jaiswal, Peter A. Beerel\\
Universiy of Southern California, Los Angeles, USA\\
\tt\small\{gdatta, liuzeyu, zihanyin, linyusun, akhilesh, pabeerel\}@usc.edu
}
\maketitle

\begin{abstract}
\vspace{-4mm}
Current computer vision (CV) systems use an image signal processing (ISP) unit to convert the high resolution raw images captured by image sensors to visually pleasing RGB images. Typically, CV models are trained on these RGB images and have yielded state-of-the-art (SOTA) performance on a wide range of complex vision tasks, such as object detection. In addition, 
in order to deploy these models on resource-constrained low-power devices, recent works have proposed in-sensor and in-pixel computing approaches that try to partly/fully bypass the ISP and yield significant bandwidth reduction between the image sensor and the CV processing unit by downsampling the activation maps in the initial convolutional neural network (CNN) layers. However, direct inference on the raw images degrades the test accuracy due to the difference in covariance of the raw images captured by the image sensors compared to the ISP-processed images used for training. Moreover, it is difficult to train deep CV models on raw images, because most (if not all) large-scale open-source datasets consist of RGB images. To mitigate this concern, we propose to invert the ISP pipeline, which can convert the RGB images of any dataset to its raw counterparts, and enable model training on raw images. 
We release the raw version of the COCO dataset, a large-scale benchmark for generic high-level vision tasks. For ISP-less CV systems, training on these raw images result in a ${\sim}7.1\%$ increase in test accuracy on the visual wake works (VWW) dataset compared to relying on training with traditional ISP-processed RGB datasets. To further improve the accuracy of ISP-less CV models and to increase the energy and bandwidth benefits obtained by in-sensor/in-pixel computing, we propose an energy-efficient form of analog in-pixel demosaicing that may be coupled with in-pixel CNN computations. 
When evaluated on raw images captured by real sensors 
from the PASCALRAW dataset, our approach results in a $8.1\%$ increase in mAP. 
Lastly, we demonstrate a further $20.5\%$ increase in mAP by using a novel application of few-shot learning with thirty shots each for the novel PASCALRAW dataset, constituting 3 classes.

\end{abstract}

\section{Introduction}
\label{sec:intro}

\begin{figure}
\centerline{\includegraphics[scale=0.21]{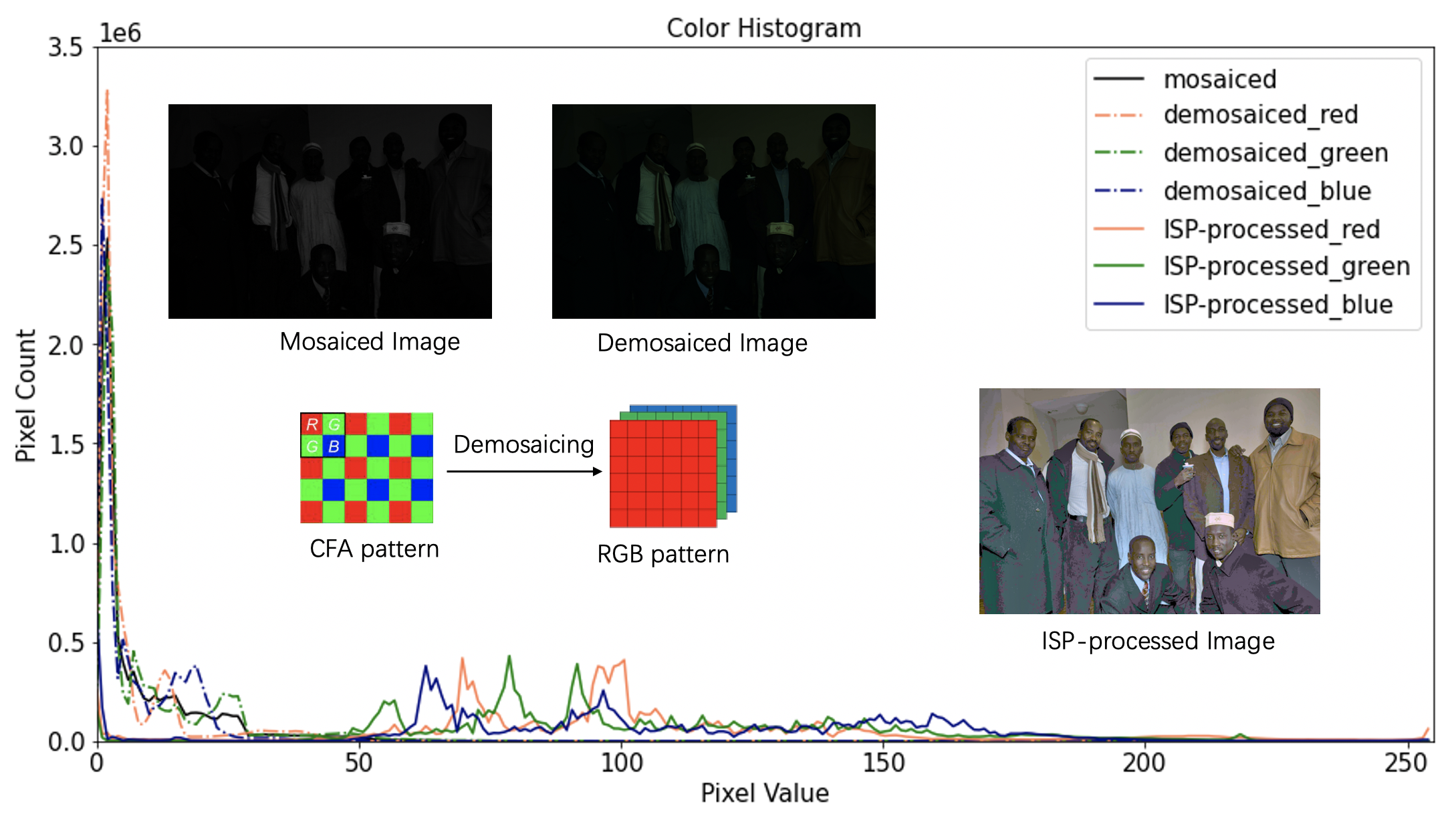}}
\caption{
 Difference in frequency distributions of pixel intensities between mosaiced raw, demosaiced, and ISP-processed images.}
\label{fig:demosaic}
\end{figure}

Modern high-resolution cameras generate huge amount of visual data arranged in the form of raw Bayer color filter arrays (CFA), also known as a mosaic pattern, as shown in Fig. \ref{fig:demosaic}, that need to be processed for downstream CV tasks \cite{surveillance, sensor_increasing}. An ISP unit, consisting of several pipelined processing stages, is typically used before the CV processing to convert the raw mosaiced images to RGB counterparts \cite{isp4ml, isp_intel, lubana_isp, isp_opt_cvpr20}. The ISP step that converts these single-channel CFA images to three-channel RGB images is called demosaicing. Historically, ISP has been proven to be extremely effective for computational photography applications, where the goal is to generate images that are aesthetically pleasing to the human eye \cite{isp_opt_cvpr20, isp2}. 
However, is it important for high-level CV applications, such as face detection by smart security cameras, where the sensor data is unlikely to be viewed by any human? Existing works \cite{isp_intel, isp4ml, lubana_isp} show that most ISP steps can be discarded with a small drop in the test accuracy for large-scale image recognition tasks.
The removal of the ISP can potentially enable existing in-sensor \cite{pinkhan2021jetcas, chen2020pns, sony2020vision} and in-pixel \cite{scamp2020eccv,Mennel2020UltrafastMV, datta2022scireports, datta2022hsipip, p2m-detrack} computing paradigms to process CV computations, such as CNNs partly in the sensor, and reduce the bandwidth and energy incurred in the data transfer between the sensor and the CV system.
Moreover, most low-power cameras with a few MPixels resolution, do not have an on-board ISP \cite{TI2020vision}, thereby requiring the ISP to be implemented off-chip, increasing the energy consumption of the total CV system. 

Although the ISP removal can facilitate model deployments in resource-constrained edge devices, one key challenge is that most large-scale datasets, that are used to train CV models, are ISP-processed.
Since there is a large co-variance shift between the raw and RGB images (please see Fig. \ref{fig:demosaic} where we show the histogram of the pixel intensity distributions of RGB and raw images), models trained on ISP-processed RGB images and inferred on raw images, thereby removing the ISP, exhibit a significant drop in the accuracy. 
One recent work has leveraged trainable flow-based invertible neural networks \cite{invertible_isp} to convert raw to RGB images and vice-versa using open-source ISP datasets. These networks have recently yielded SOTA test performance in photographic tasks, which we propose to modify to invert the ISP pipeline, and build the raw version of any large-scale ISP processed database for high-level vision applications, such as object detection. This raw dataset can then be used to train CV models that can be efficiently deployed on low-power edge devices without any of the ISP steps, including demosaicing. To further improve the performance of these ISP-less models, we propose a novel hardware-software co-design approach, where a form of demosaicing is applied on the raw mosaiced images inside the pixel array using analog summation during the pixel read-out operation, i.e., without a dedicated ISP unit. Our models trained on this demosaiced version of the visual wake words (VWW) lead to a $8.2\%$ increase in the test accuracy compared to standard training on RGB images and inference on raw images (to simulate the ISP removal and the in-pixel/in-sensor implementation). Even compared to standard RGB training and inference, our models yield $0.7\%$ ($1.6\%$) higher accuracy (mAP) on the VWW (COCO) dataset. Lastly, we propose a novel application of few-shot learning to improve the accuracy of real raw images captured directly by a camera (which has limited number of annotations) with our generated raw images constituting the {base} dataset.

The key contributions of our paper can be summarized as follows.
\vspace*{-0.75em}
\begin{itemize}
    \setlength\itemsep{-0.5em}
    \item Inspired by the energy and bandwidth benefits obtained by in-sensor computing approaches and the removal of most ISP steps in a CV pipeline, we present and release a large-scale raw image database that can be used to train accurate CV models for low-power ISP-less edge deployments. This dataset is generated by reversing the entire ISP pipeline using the recently proposed flow-based invertible neural networks and custom mosaicing. We demonstrate the utility of this dataset to train ISP-less CV models with raw images. 
    \item To improve the accuracy obtained with raw images, we propose a low-overhead form of in-pixel demosaicing that can be implemented directly on the pixel array alongside other CV computations enabled by recent paradigms of in-pixel/in-sensor computing approaches and that also reduces the data bandwidth.
    \item We present a thorough evaluation of our approach with both \textit{simulated} (our released dataset) and \textit{real} (captured by a real camera) raw images, for a diverse range of use-cases with different memory/compute budgets.
    \item To improve the accuracy of real raw images, 
    we propose a novel application of few-shot learning, with the simulated raw images having a large number of labelled classes constituting the {base} dataset.
\end{itemize}


\begin{figure*}[htbp]
\centerline{\includegraphics[scale=0.42]{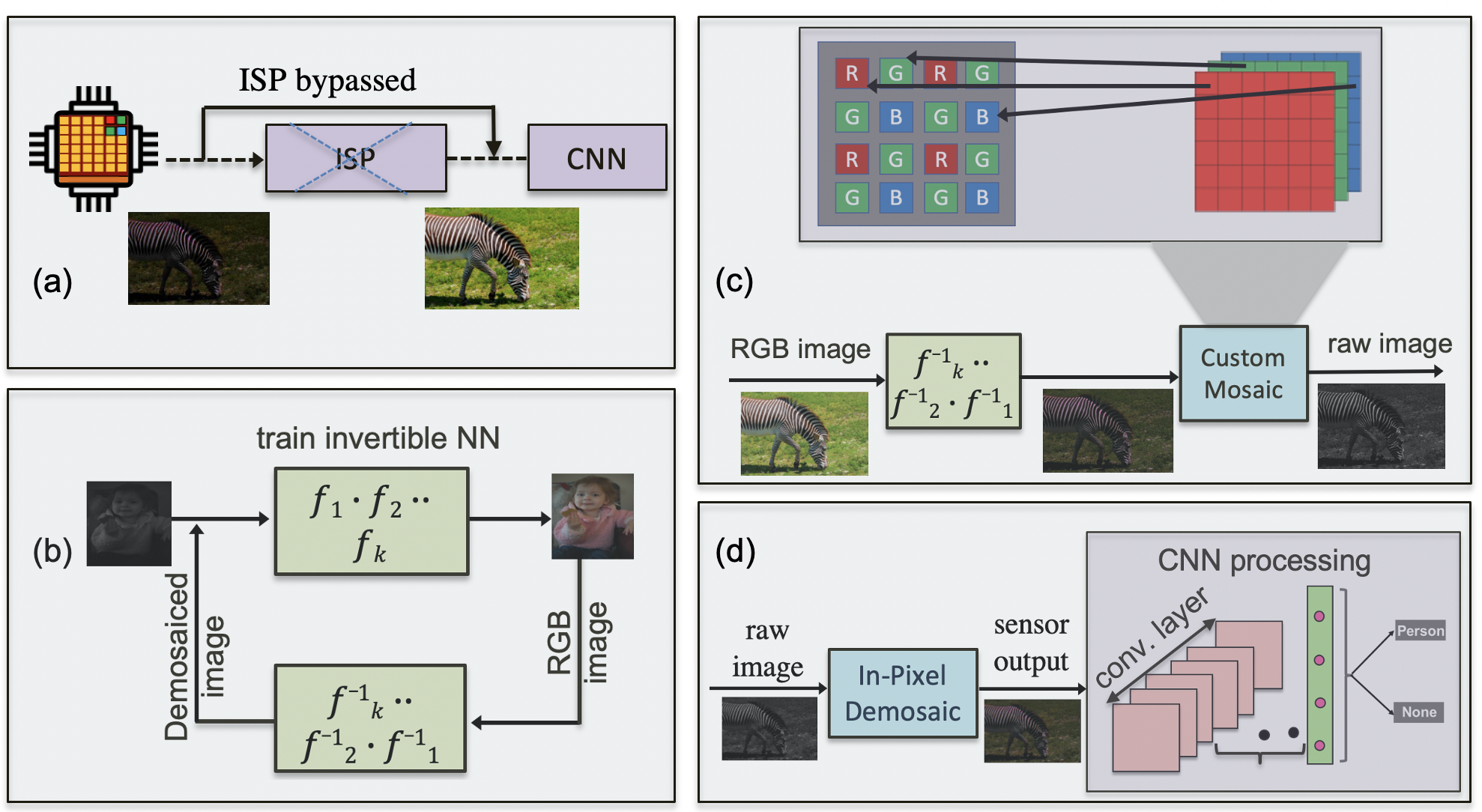}}
\caption{
 (a) Proposed ISP-less CV system, (b) Invertible NN training on demosaiced raw image, without any white balance or gamma correction, (c) Generation of raw images using the trained inverse network and custom mosaicing, and (d) Application of in-pixel demosaicing and training of the ISP-less CV models. Note the \textit{In-Pixel Demosaic} implementation in the pixel array is illustrated in Fig. \ref{fig:pixel}.}
\label{fig:training_framework}
\end{figure*}

\section{Related Works}

\subsection{ISP Reversal \& Removal}\label{subsec:isp_removal}

Since most ISP steps are irreversible, and depend on the camera manufacturer's proprietary color profile \cite{brooks2018}, it is difficult to invert the ISP pipeline. To mitigate this challenge, a few recent works \cite{Liu_2020_CVPR, ISP_compressor, Zamir2020CycleISP} proposed learning-based methods, but they result in large losses and the recovered RAW images may be significantly different from the originals captured by the camera. To reduce this loss, a more recent work \cite{invertible_isp} used a stack of $k$ invertible and bijective functions $f=f_1\cdot f_2\cdot..f_k$ to invert the ISP pipeline. For a raw input $x$, the RGB output $y$ and the inverted raw input $x$ is computed as $y=f_1\odot f_2\odot.. f_k(x)$ and $x=f^{-1}_k\odot f^{-1}_{k-1}\odot.. f^{-1}_1(y)$.

The bijective function $f_i$
is implemented through affine
coupling layers \cite{invertible_isp}. In each affine coupling layer, given a $D$ dimensional input $m$ and $d{<}D$, the output $n$ is
\begin{small}
 \begin{align}
 n_{1:d}&=m_{1:d}+r(m_{d+1:D})  \\
 n_{d+1:D}&=m_{d+1:D}\odot exp(s(m_{1:d}))+t(m_{1:d})
 \label{eq:x_isp}
 \end{align}
 \end{small}
where $s$ and $t$ represent scale and translation functions from
$R^{d}$ to $R^{D-d}$ that are realized by neural networks, $\odot$ represents the Hadamard product, and $r$ represents an arbitrary function from $R^{D-d}$ to $R^{d}$. The inverse step is
\begin{small}
\begin{align}
m_{d+1:D}&=(n_{d+1:D}-t(n_{1:d}))\odot exp(-s(n_{1:d})) \\
m_{1:d}&=n_{1:d}-r(m_{d+1:D})
 \label{eq:x_isp}
 \end{align}
 \end{small}
 
\noindent
The authors then utilize invertible $1{\times}1$ convolution, proposed in \cite{NEURIPS2018_d139db6a}, as the learnable permutation function to revert the channel order for the subsequent affine coupling layer. 

Recent works have also investigated the role of the ISP in image classification and the impact of its' removal/trimming on accuracy for energy and bandwidth benefits. For example, \cite{isp4ml} demonstrated that removal of the whole ISP during edge inference results in a ${\sim}8.6\%$ loss in accuracy with MobileNets \cite{sandler2018mobilenetv2} on ImageNet \cite{imagenet}, which can mostly be recovered by using just the tone-mapping stage. Another work \cite{isp_intel} attempted to integrate the ISP and CV processing using tone mapping and feature-aware downscaling blocks that reduce both the number of bits per pixel and the number of pixels
per frame. A more recent work \cite{ISP_distill} used knowledge distillation on an ISP neural network model to align the logit predictions of an off-the-shelf pretrained model for raw images with that for ISP-processed RGB images.

\subsection{Few-Shot Object Detection}\label{subsec:few_shot_det}

In recent years, few-shot object detection (FSOD) has gained significant traction as ML accuracy in low-data scenarios continues to improve. There are two mainstream training paradigms in FSOD, meta-learning and finetune-based methods. 
Meta-learning methods attempt to capture aggregated information from multiple annotated data-rich support datasets. Thus, when required to train on a dataset with novel classes and less data, the model can leverage the prior knowledge learned from support datasets to generalize to new classes. For example, \cite{Kang_2019_ICCV} used a re-weighting module to adjust coefficients of the query image meta features by capturing global features of the support images to be suitable for novel object detection. Authors in \cite{yan2019meta} proposed a Predictor-head Remodeling Network (PRN) module to generate class-attentive vectors to provide aggregated features between support and query images for the meta-learner predictor head. Additionally, \cite{fan2020few} introduced an attention-based region proposal network to match the candidate proposals with the support images and a multi-relation detector which can measure the similarity between proposal boxes from the query and the support objects.
Compared to meta-learning which requires a complicated training process, the finetune-based methods have a simpler pipeline. For instance, \cite{wang2020few} proposed the two-stage fine-tune based approach (TFA) which only finetunes the bounding box classification and regression parts on a class-balanced training set, 
but outperforms many meta-learning methods. Moreover, to mitigate misclassifying novel instances as confusing base classes, \cite{sun2021fsce} introduced contrastive learning into the FSOD pipeline, that helps the learned target features represent high intra-class similarity and inter-class variability. 
\vspace{-3mm}

\section{Inverting ISP Pipeline}

Similar to \cite{invertible_isp}, we propose to generate the raw demosaiced images from ISP-processed RGB images using the affine coupling layers described in Section \ref{subsec:isp_removal}. However, \cite{invertible_isp} models the ISP pipeline on the demosaiced, white-balanced and gamma-corrected raw image, and hence, the invertible ISP pipeline does not generate the raw image that are captured directly by a camera. The authors apply gamma correction on the RAW data (i.e. without storing on disk) to compress the dynamic range for faster convergence. Hence, for ISP-less in-sensor CV systems, the naive application of the invertible ISP pipeline proposed in \cite{invertible_isp} will require performing these operations in the sensor.
This is challenging due to the limited compute/memory footprint available in the pixel array and the periphery. In particular, traditional demosaicing involves matrix operations that involve interpolation (nearest neighbour, bi-linear, bi-cubic, etc.) techniques which scale with the input resolution. Moreover, white balancing involves a variable gain amplification for each pixel location which requires complex control logic, and gamma correction involves logarithmic computation which is challenging to process using analog logic in advanced high-density pixels. 

For these reasons, we propose to train an invertible network on the demosaiced images from the MIT-Adobe 5K dataset \cite{mit-adobe_5k}. Despite our focus on classification/detection tasks, we propose to use this photographic dataset to train the invertible ISP because we do not have large-scale ground truth raw-RGB image pairs for those tasks. We train using demosaiced images because the input size of the invertible neural network must be equal to its output size. Once trained, we use this network to obtain the raw demosaiced images from the ISP-processed RGB images from the large-scale classification/detection datasets. 

We then invert the demosaicing i.e., perform the mosaicing operation by selecting the appropriate pixel color corresponding to each location, as shown in Fig. \ref{fig:training_framework}. For example, to generate the red pixel in a particular mosaiced RGGB patch, we select the pixel intensity of the red channel in the same location as in the demosaiced image. 
Although this final mosaiced image is obtained after inverting the entire ISP pipeline, it might still be slightly different from the raw image captured by a camera. This is partially because we do not explicitly model the latent distribution of the different ISP steps to stabilize the training of the invertible network. We mitigate this concern using few-shot learning. 

\begin{figure*}[htbp]
\centering
\includegraphics[width = 0.8\textwidth]{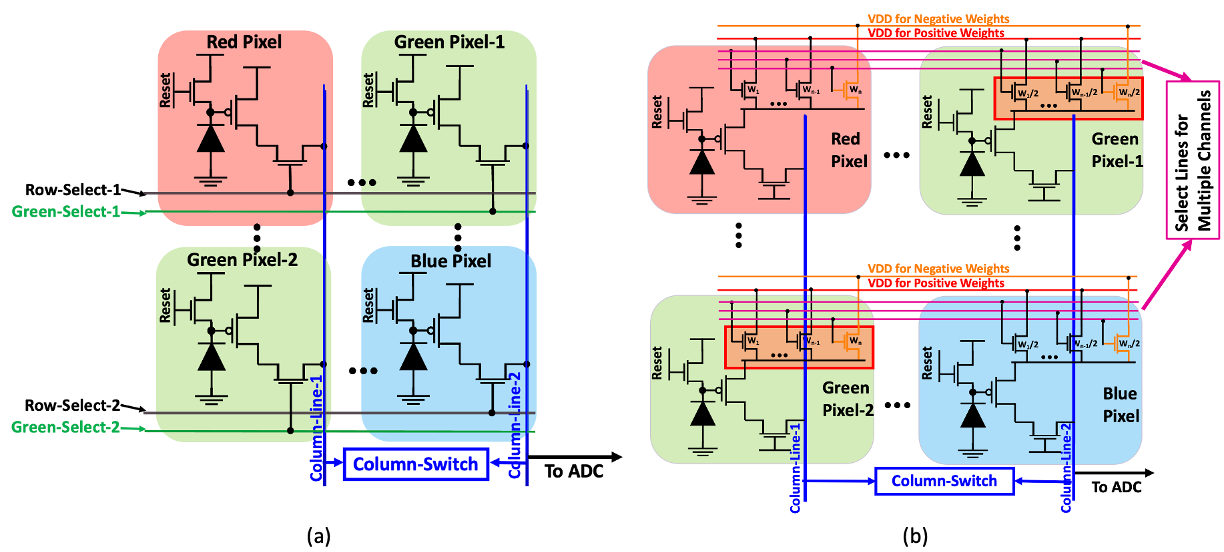}
\vspace{-1mm}
\label{sum}
\caption{Implementation of the proposed (a) demosacing and (b) demosaicing coupled with in-pixel convolution for ISP-less CV.}
\vspace{-5mm}
\label{fig:pixel}
\end{figure*}


\section{Requirement for Demosaicing}

Although training on raw images in the Bayer CFA format increases the test accuracy of ISP-less CV applications, it might lack the representation capacity that multiple colors spanning different spectral bands might provide for each pixel location. Hence, a natural question is can we increase this capacity without an additional ISP unit? Since demosaicing is the ISP technique that yields separate RGB channels from the raw CFA format, one intuitive idea is to implement demosaicing directly in the pixel array, and then process the CV computations required for CNNs using in-pixel/in-sensor computing. However, as explained above, traditional demosaicing approaches involve complex operations which are hard to map on the pixel array, especially when the pixel array needs to process the initial CNN layers in the in-pixel computing paradigms. Hence, we propose a low-overhead custom in-pixel demosaicing approach that significantly increases the test accuracy for our benchmarks compared to inference on raw images.

\section{Proposed Demosaicing Technique}\label{subsec:demosaicing}

We propose to implement a simple but effective custom demosaicing operation inside the analog pixel array.  
Let us consider a demosaiced RGB image with shape $X{\times}Y{\times}3$, to be processed for CV applications. Then, our custom demosaicing technique requires the input mosaiced raw image to have a shape $2X{\times}2Y$, such that each $2{\times}2$ RGGB patch produces the corresponding $3$ channels for a single pixel, thereby yielding a $25\%$ reduction in data dimensionality. Functionally, the custom demoisaicing copies the red and blue pixel intensities from the camera to the demosaiced RGB channel output, while the two green pixels from the RGGB patch of the camera pixel array are \textit{averaged} to produce one effective value for green pixel intensity. While the summation is performed by analog computation inside the pixel array described below, the division is performed in the digital domain after the analog to digital converter (ADC) in the pixel periphery by a simple logical right shift operation. 

The proposed implementation of the pixel array to accomplish this custom demosaicing functionality is shown in Fig. \ref{fig:pixel}(a). We propose to include two select lines for each row of the pixel array - the first set of select lines called `Row-Select' are connected to the select transistors of the red and blue pixels, while the second set of select lines called `Green-Select' are connected only to the green pixels. Essentially, the pixels in RGGB Bayer pattern are connected in an interleaved manner to the two select lines. Therefore, the read-out of the red and blue pixels are controlled by the `Row-Select' lines, while the `Green-Select' lines control the read-out of the green pixels. Now consider activating two rows of `Row-Select' lines (Row-Select-1 and Row-Select-2) in the 2x2 pixel array of Fig. \ref{fig:pixel}(a). This would result in read-out of the red and blue pixels on  `Column-Line-1' and `Column-Line-2, respectively. The two green pixels would remain deactivated as the `Green-Select' lines are kept at low voltage. In a subsequent cycle, the two `Row-Select' lines are kept at low voltage and the two `Green-Select' lines are activated by pulling them to high voltage. Further, the two `Column-Lines' are connected together by closing the `Column-Switch', shown in Fig. \ref{fig:pixel}(a). Consequently, the voltage on the now connected `Column-Lines' represents the accumulated response of the two green pixels, which are fed to column-ADCs for analog to digital conversion. Note, the proposed scheme is similar to pixel binning approaches \cite{bock2008methods,bin1, bin2, bin3}, except that in this case binning is selectively performed only for the two green pixels in each patch of Bayer RGGB pattern using interleaved connections to `Row-Select' and `Green-Select' lines. In summary, in two cycles, wherein two rows of `Row-Select' and `Green-Select' lines are activated in each cycle, the proposed scheme can generate demosaiced red, blue, and green pixels. Note, since we are able to read two rows (consisting of RGGB pixels) in two cycles, the proposed scheme does not incur any overhead in terms of read-out speed (or frame-rate) of the camera.

In yet another approach, we propose to combine the custom demoisacing and the computations of the first-layer of CNN inside pixel array using the P$^2$M (Processing-in-pixel-in-memory) paradigm proposed in \cite{datta2022scireports}, as shown in Fig. \ref{fig:pixel}(b). 
Modifying the P$^2$M pixel array of \cite{datta2022scireports}, Fig. \ref{fig:pixel}(b) presents a novel pixel array that can combine demosaicing and convolution computations using memory-embedded pixels. Essentially, the CNN weights associated with two green pixels in a single patch of Bayer RGGB pattern are kept the same. This is achieved by keeping the sizes of the weight transistors the same across the two green pixels. Further, these transistor weights are kept such that each set of weight transistors for a single pixel has half of the effective algorithmic weight value associated with the green channel in the input convolutional layer. This ensures that the resultant analog dot product obtained from the P$^2$M scheme \cite{datta2022scireports} involve effective averaging of the intensities of the green pixels and then multiplying it with the corresponding weights associated with the convolutional layer.

While the in-pixel convolution on the demosaiced image can lead to significantly higher bandwidth reduction \cite{datta2022scireports} (quantified later in Section \ref{subsec:BW_benefits}), the analog non-idealities involved in the multiply-and-accumulate operation and weight mismatch in the green pixels can lead to large errors, require re-training the entire CNN network, and introduce manufacturing challenges, which might require non-trivial changes to the design pipeline of sensors.



\section{Few-Shot Learning}

Compared to the abundance of RGB image datasets, it is difficult to obtain large-scale annotated raw images. For example, to the best of our knowledge, the only raw image database for classification/detection tasks, PASCALRAW, contains only 4,259 annotated images, with 3 object classes, which are not enough to train a deep CV model. Even with a pre-trained model on a large-scale RGB dataset, it might be difficult to fine-tune on this small-scale raw dataset (due to the co-variance shift) and yield satisfactory performance.

As described in Section \ref{subsec:few_shot_det}, recent works have proposed a plethora of few-shot learning approaches that achieve great performance on datasets with some novel classes, and a few images per class.
Our problem is not exactly same as a typical few-shot learning setup, given we can find a large-scale annotated RGB image database, having the same classes as the raw dataset. For example, the Microsoft COCO dataset consists of 80 classes, which can cover objects from a range of applications such as autonomous driving, aerial imagery recognition, and can be used for fine-tuning on a raw image database with a subset of their classes. We propose to leverage TFA \cite{wang2020few} (see Section \ref{subsec:few_shot_det}) for this fine-tuning process, and to the best of our knowledge, this is the first application of few-shot learning in improving the accuracy/mAP of raw images.

However, naively applying TFA with COCO as the base dataset can only bring limited improvement in accuracy, due to the co-variance shift between the RGB and raw images. Note that in typical few-shot learning setups, such as TFA, the images in the base and novel datasets are assumed to have similar intensity distributions \cite{wang2020few}. Hence, we propose a novel application of few-shot learning which leverages our simulated raw COCO dataset as the base class to increase the test mAP on the real raw dataset. 
We choose a class-balanced subset of the real raw dataset as the samples with `novel class' to perform TFA on the model pretrained on our raw COCO dataset to further improve the mAP.


\section{Experimental Setup}
\label{sec:experiments}



\subsection{Implementation Details}
 We evaluate our proposed method on three CNN backbones/frameworks with varying complexities and use-cases as described below. For object detection experiments, we use the \cite{mmdetection} framework, while for few-shot learning, we use the \textit{mmfewshot} \cite{mmfewshot2021} and \textit{FsDet} \cite{wang2020few} framework. Our training details are provided in the supplementary materials. \\
\textit{MobileNetV2}~\cite{sandler2018mobilenetv2}: A lightweight depthwise convolution neural network that has gained significant traction for being deployed on resource-constrained edge devices, such as mobile devices. In this work, we use a lower complexity version of MobileNetV2, namely MobileNetV2-0.35x \cite{NEURIPS2020_ebd9629f}, which shrinks the output channel count by 0.35$\times$ to satisy the compute budget of 60M floating point operations (FLOPs) representing standard micro-controllers, where ISP-less CV may be the most relevant. \\
 \textit{Faster R-CNN}~\cite{ren2015faster}: A two-stage object detection framework that consists of a feature extraction, a region proposal, and a RoI pooling module. For our experiments, we use ResNet101 as the backbone network for feature extraction, since MobileNetV2 significantly degrades the test mAP compared to SOTA.  \\
\textit{YOLOv3} \cite{redmon2018yolov3}: YOLOv3 (You Only Look Once, Version 3) is a real-time object detection framework that identifies specific objects in videos or images. We use MobileNetV2 as the backbone network for feature extraction in YOLOv3.



\subsection{Dataset Details}
 
We evaluate our proposed approaches on the simulated raw versions of the Visual Wake Words (VWW) and COCO datasets, and a real raw dataset captured by a real camera introduced in \cite{pascal-voc-2012}. The details of the datasets are below. \\
\noindent
\textit{VWW}~\cite{chowdhery2019visual}: The Visual Wake Words (VWW) dataset 
consists of high resolution images that include visual cues to “wake-up" AI-powered resource-constrained home assistant devices that require real-time inference. 
The goal of the VWW challenge is to detect the presence of a human in the frame (a binary classification task with 2 labels) with very little resources-close to $250$KB peak RAM usage and model size, which is only satisfied by MobileNetV2-0.35x, and hence, used in our experiments. \\
\textit{Microsoft COCO}: To evaluate on the multi-object detection task, we use the popular Microsoft COCO dataset~\cite{coco}. 
Specifically, we use an image resolution of $1333{\times}800$ for the Faster-RCNN framework, and $416{\times}416$ for the YoloV3 framework \cite{redmon2018yolov3}, the same as used in~\cite{redmon2018yolov3}. We use the 80 available classes used for our experiments.
 We evaluate the performance of each method using mAP averaged for IoU $\in\{0.5, 0.75, [0.5:0.05:0.95]\}$, denoted as mAP@0.5, mAP@0.75 and mAP@[0.5, 0.95], respectively. Note that we also report the individual mAPs for small (area ${<}32^2$ pixels), medium (area between $32^2$ and $96^2$ pixels), and large (area ${>}96^2$ pixels) objects. \\
\textit{PASCALRAW}: This RAW image database was developed to simulate the effect of algorithmic hardware implementations such as embedded feature extraction at the image sensor or readout level on end-to-end object detection performance. 
The annotations of this dataset were made in accordance with the original PASCAL VOC guidelines \cite{pascal-voc-2012}. 
For the few-shot learning experiments, we choose 29 images containing the class `bicycle', 25 images containing the class `car' and 21 images containing the class `person' to construct a balanced training set where each class has 30 annotated objects (i.e., 30-shot), and use the remaining 4178 images as the test dataset.  

\section{Experimental Results}
\subsection{VWW Results}

For VWW, we compare the accuracy of the tinyML-based MobileNetV2-0.35x model with our proposed demosaicing and in-pixel computing technique against inference on mosaiced raw and RGB images in Table \ref{tab: comparison on vww}. We also compare our approach with traditional demosaicing (\textit{opencv} library in Python), white balancing (\textit{rawpy} package in Python), and gamma correction. 
Note that, as shown in Table \ref{tab: comparison on vww}, using identically distributed images during training yields the best accuracy during inference. 

Table \ref{tab: comparison on vww} further illustrates that using  the off-the-shelf model pre-trained on ISP-processed images yields an accuracy of $81.97\%$ when deployed on an ISP-less CV system with our in-pixel demosaicing, which is $7.32\%$ lower compared to the ISP-processed inference. Note that we cannot avoid the demosaicing step, because the pre-trained model is trained with 3 channel input images. 
With the generated mosaiced image database from our invertible pipeline, the accuracy gap (training and testing both on the mosaiced image) reduces to $2.82\%$. Additionally, with our in-pixel demosaicing on this mosaiced image, we yield an accuracy of $89.92\%$, which is even $0.63\%$ higher than the RGB test accuracy. Appending the first layer convolution inside the pixel, coupled with the demosacing results in a little lower accuracy of $89.07\%$. 

\begin{table}
\caption{Evaluation of our approach on ISP-less CV systems with MobileNetV2-0.35x on VWW dataset.  Demosaiced$^1$ denotes traditional demosaicing, while demosaiced$^2$ denotes our in-pixel demosaicing. WB, GC, and IPC denotes white balance, gamma correction, and in-pixel computing. Also, note that models trained on mosaiced images can only be tested with mosaiced images.}
\small
\vspace{-4mm}
\label{tab: comparison on vww}
\begin{center}
\setlength{\tabcolsep}{0.5mm}
{
\begin{threeparttable}
\begin{tabular}{l|c|c|c}
\hline
\hline
Method & \multicolumn{3}{c}{Test Acc. (\%)} \\
\hline

\diagbox[]{Training}{Inference} 
  & Mosaiced & demosaiced$^2$ & IPC  \\
\hline

Mosaiced
    & 87.47 & - & -  \\
demosaiced$^1$
    & - & 88.84 & 88.04  \\
demosaiced$^2$
    & - &  \textbf{89.92} & \textbf{89.07}  \\
demosaiced$^1$+WB
    & - & 86.47 & 86.23   \\
demosaiced$^1$+WB+GC
  & - & 82.70  & 81.45  \\
ISP-processed
 & - & 81.97 & 81.43  \\
\hline
\end{tabular}
\end{threeparttable}
}
\vspace{-5mm}
\end{center}
\end{table}

\subsection{COCO raw Results}
\begin{table}
\caption{mAP on different versions of the COCO raw dataset to emulate ISP-less CV systems using a Faster R-CNN framework with ResNet101 backbone.}
\small
\centering
\setlength{\tabcolsep}{1.6mm}
{
\begin{threeparttable}
\begin{tabular}{lcccccc}
\toprule
{} &  \multicolumn{6}{c}{mean average precision}   \\
model &                             0.5:0.95&   0.5&    0.75&   S&      M&      L  \\
\midrule
baseline&                               33.8&    50.5&    37.0&    16.6&    36.6&    46.7\\
demosaiced$^1$&                               \textbf{42.8}&    \textbf{64.1} &   \textbf{47.1} &    \textbf{25.6} &    46.9&   \textbf{55.0}\\ 
mosaiced&                           29.4&    45.7&   31.8&    12.7&    32.1&    42.9\\
demosaiced$^2$&                           37.8 &   57.7&   39.8&   20.2&    \textbf{48.6} &   53.2\\
\bottomrule
\end{tabular}
\begin{tablenotes}    
        \footnotesize   
        \item[1] `baseline' indicates testing on our proposed COCO raw datset with model pretrained on ISP-processed COCO dataset
        \item[2] `demosaiced$^1$' indicates training and testing on our proposed COCO raw dataset 
        \item[3] `mosaiced' indicates training and testing on mosaiced images obtained from the COCO raw dataset from our invertible ISP
        \item[4] `demosaiced$^2$' means training and testing on our in-pixel demosaiced images
      \end{tablenotes}          
    \end{threeparttable} 
    }
    \vspace{-3mm}
\label{results/few-shot-COCO}
\end{table}

\begin{table*}[t]
\caption{Comparison of our proposed approach on PASCALRAW dataset.}
\small
\label{tab: comparison on pascal raw}
\begin{center}
{
\begin{threeparttable}
\begin{tabular}{l|c|c|c|c|c|c|c}
\hline
\hline

\multirow{2}{*}{Framework} & \multirow{2}{*}{Method} & \multicolumn{6}{c}{mAP} \\
\cline{3-8}
& &  @[0.5,0.95] &@0.5 & @0.75 & @small & @medium & @large \\
\hline

\multirow{6}{*}{Yolov3}

&ISP-processed&                               2.7&    8.2&    1.2&    0.2&    2.2&    4.4\\
&ISP-processed+few-shot$^{*}$&                5.2& 15.4& 2.4& 0.6& 5.5& 7.9           \\
&ISP-processed+few-shot$^{**}$&                6.2& 17.0& 3.3& 0.2& 3.9& 11.2   \\
&demosaiced raw&                               13.4&    38.5&   5.4&    0.9&    12.2&   22.9\\ 
&demosaiced raw+few-shot$^{*}$&                         16.9& 40.6& 10.9& 0.5& \textbf{17.8}& 26.3 \\
&demosaiced raw+few-shot$^{**}$&                           \textbf{20.8}&    \textbf{47.4}&   \textbf{14.5}&    \textbf{0.9}&    17.3&    \textbf{30.4}\\

\hline

\multirow{6}{*}{Faster RCNN} 
& ISP-processed&                               1.2&    4.2&    0.2&    0.0&    1.3&    3.5\\
&ISP-processed+few-shot$^{*}$&                           5.9&   14.8&    3.3&    0.0&    3.8&    8.6\\
&ISP-processed+few-shot$^{**}$&                           9.5&    26.0&   4.2&    0.0&    6.6&    15.0\\
&demosaiced raw&                               9.3&    29.9&   2.2&    1.7&    10.5&   19.5\\
&demosaiced raw+few-shot$^{*}$&                          27.4&   52.8&   25.7&   6.9&     27.1&  37.3\\
&demosaiced raw+few-shot$^{**}$&               \textbf{29.8}&   \textbf{58.1}&   \textbf{28.0}&   \textbf{8.0}&    \textbf{28.1}&   \textbf{40.6}\\

\hline
\hline
\end{tabular}
\begin{tablenotes}
    \footnotesize{
        \item * The experiments apply few-shot learning with 30 shots of both base classes and novel classes.   
        \item ** The experiments apply few-shot learning with 30 shots of only the novel classes.
    }
\end{tablenotes}
\end{threeparttable}
}
\vspace{-5mm}
\end{center}
\end{table*}
The detailed results on COCO raw dataset are summarized in Table \ref{results/few-shot-COCO}. Our experiments indicate that direct inference on the COCO demosaiced raw dataset using the model pre-trained on COCO ISP-processed RGB dataset yields an mAP of $33.8\%$, which is $7.2\%$ lower compared to ISP-processed inference. 
Note that the mAP of small objects reduces significantly by nearly $35\%$. However, with finetuning on our COCO demosaiced raw dataset, the mAP increases to $42.8\%$. 
Unlike in VWW, where models can be accurately trained from scratch, training and testing on the COCO mosaiced raw image leads to a reduced mAP of $29.4\%$. This reduction might be because the pre-trained model (where the backbone is also pre-trained on ImageNet) cannot be leveraged because of the difference in the number of input channels. Lastly, applying our proposed in-pixel demosaicing on the mosaiced raw dataset yields an mAP of $37.0\%$, which is $5.0\%$ lower than ISP-processed inference, unlike VWW. This might be because our demosaicing reduces the spatial resolution of the image, which might be detrimental for the complex object detection task. Interestingly, our approach is effective in detecting medium sized objects, and achieves the highest mAP of $48.6\%$.

\subsection{PASCALRAW Results}
\subsubsection{YOLOv3}
Table \ref{tab: comparison on pascal raw} shows the performance of six different methods with YOLOv3 on the PASCALRAW dataset. Direct inference on this dataset with models pre-trained on ISP-processed COCO dataset yields only $2.7\%$ mAP due to the significant co-variance shift between the two datasets. Using the ISP-processed base dataset, we compare two different few-shot learning approaches, one where we use 30 shots for both the base and novel classes, and the other where we use 30 shots only for the novel classes. We observe the latter leads to $1.0\%$ higher mAP compared to the former, which might be because the former may underfit to the three target classes due to its improved generalization. Note, due to the difference in the dataset distributions, few-shot learning fails to significantly increase the mAP, as observed from the modest mAP improvement from $2.7\%$ to $6.2\%$. On the other hand, after finetuning on our custom demosaiced COCO raw dataset (without any few-shot learning), the mAP increases by more than $4\times$ to $13.4\%$. This strongly demonstrates the effectiveness of our large-scale raw database. Lastly, applying few-shot learning with this base raw dataset further increases the mAP to $20.8\%$. 

\vspace{-3mm}
\subsubsection{Faster R-CNN}
 We perform a series of similar experiments with Faster R-CNN model with ResNet101 backbone on the PASCALRAW dataset. As we can see in Table \ref{tab: comparison on pascal raw}, the results are consistent with that from the YOLOv3 model, except that there is no mAP increase with fine-tuning on the COCO raw dataset compared to applying few-shot learning with the ISP-processed base dataset. This might be because our demosaicing approach that incurs $4\times$ spatial down-sampling might not be that competitive compared to the faster R-CNN framework with ultra-high input resolution. 
 Applying few-shot learning with 30 shots of only the novel classes on our custom demosaiced COCO raw dataset yields an mAP of $29.8\%$, which is $28.6\%$ higher compared to directly using the model pretrained on the ISP-processed COCO dataset.

\begin{figure*}
\centering
\includegraphics[width = 0.96\textwidth]{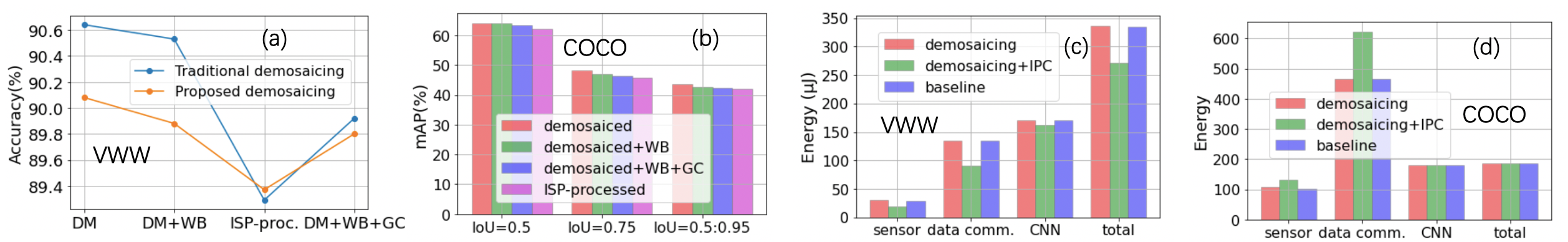}
\vspace{-1mm}
\label{sum}
\caption{Comparison of the (a) accuracy and (b) mAP of our proposed demosaicing method with different ISP pipelines on COCO dataset with Faster-RCNN framework with ResNet101 backbone and VWW dataset with MobileNetV2-0.35x respectively, where DM denotes our proposed demosaicing technique, WB and GC denote white balancing and gamma correction respectively. The energy consumptions of our approaches are compared with the normal pixel read-out in (c) and (d) on VWW and COCO respectively, where IPC denotes in-pixel computing. Note, for (d), the energy unit is ${\mu}J$ for `sensor' \& `data comm.', and $100{\mu}J$ for `CNN' \& `total'.}
\vspace{-5mm}
\label{fig:ablation}
\end{figure*}

\subsection{Comparison with Prior Works}

We compare the test accuracy and mAP obtained by our ISP-less CV models with existing similar works on the VWW and COCO dataset respectively in Fig. \ref{fig:ablation}(a-b). As we can see, we yield similar performance on average compared to using the invertible ISP pipeline proposed in \cite{invertible_isp}, while providing bandwidth and energy reduction quantified in Section \ref{subsec:BW_benefits}. Even compared to testing on ISP-processed RGB images which require the entire ISP pipeline, we obtain $0.63\%$ ($1.6\%$) increase in accuracy (mAP) on the VWW (COCO) dataset. It is difficult to directly compare our approach with other works \cite{isp4ml, isp_intel}, as they do not release the ISP model, and evaluate the impact of the removal of different ISP stages on in-the-wild datasets, such as ImageNet \cite{imagenet} and KITTI \cite{kitti}, which may not be a relevant use-case of ISP-less low-power edge deployment.

\subsection{Bandwidth \& Energy Benefits}\label{subsec:BW_benefits}

Removing the entire ISP pipeline, and applying the proposed in-pixel demosaicing operation directly on the raw images can lead to significant energy and bandwidth savings, thereby aiding the deployment of CNN models on ultra low-power edge devices. The complete image captured by the sensors is transmitted to a down-stream SoC processing the ISP and CV units typically through energy-hungry MIPI interfaces, which cost significant bandwidth \cite{on-sensor-tinyML}. As explained in Section \ref{subsec:demosaicing}, the demosaicing operation leads to a dimensionality reduction of $\frac{4}{3}$, which implies a $25\%$ reduction in bandwidth. Quantizing the demosaiced outputs to 8-bits using custom ADCs (inputs to modern CNNs have unsigned 8-bit representation) leads to a $(\frac{12}{8})$ or $50\%$ reduction in bandwidth, assuming the raw image has a bit-depth of $12$ \cite{onsemi:AR0135AT}. Lastly, appending the first convolution layer inside the sensor yields a $3{\times}$ increase in bandwidth for MobileNetV2-0.35x. This is convolutional layer has a stride of $2$, which implies a $4\times$ dimensionality reduction, while there is a $(\frac{8}{3})$ dimensionality increase due to the 3 channels in the input demosaiced image and 8 output channels in the first convolutional layer. In summary, the total bandwidth/data transmission energy reduction due to our proposed demosaicing operation is $75\%$, while for the in-pixel computing approach (on the proposed demosaiced image as illustrated in Section \ref{subsec:demosaicing}) is $12\times$. 

Note that this energy benefit is in addition to the energy savings obtained by removing the ISP operations in an SoC, and transferring the ISP output to a CV processing unit. It is difficult to accurately quantify this saving as it depends on the underlying hardware implementation and dataflow, as well as the proprietary implementation of ISP. That said, we compare the sensor (pixel+ADC), data communication, and the CNN energy consumption of our demosaicing and in-pixel computing approaches with normal pixel read-out in Fig. \ref{fig:ablation}(c-d). While Fig. \ref{fig:ablation}(c) represents the tinyML use-case on VWW using MobileNetV2-0.35x, Fig. \ref{fig:ablation}(d) represents the more difficult use-case on COCO using Yolov3. We compute the pixel energies using our in-house circuit simulation framework, while the ADC, data communication, and CNN energies are obtained from \cite{datta2022scireports}. While our demosaicing approach incurs a sensor energy overhead of ${\sim}5\%$ on average, the proposed in-pixel implementation reduces (increases) the sensor energy by $33\%$ (${23\%}$) on VWW (COCO) with MobileNetV2-0.35x (YoloV3). The energy increase is due to the increased number of convolutional output channels (first layer) in the MobileNet backbone of YoloV3.

\section{Discussions}

In this work, we propose an ISP-less computer vision paradigm to enable the deployment of CNN models on low-power edge devices that involve processing close to the sensor nodes with limited compute/memory footprint. Our proposal has two significant benefits: 1) We release a large-scale RAW image database that can be used to train and deploy CNNs for a wide range of vision tasks (including those related to photography) and 2) Our hardware-software co-design approach leads to significant bandwidth savings compared to traditional CV pipelines. To the best of our knowledge, this is the first work to address the widely overlooked ISP pipeline in near-sensor and in-sensor processing paradigms while also proposing novel in-pixel schemes for custom demosaicing, coupled with convolution computation. Our proposed approach increases the test accuracy (mAP) of a tinyML (generic object detection) application by $7.32\%$ ($7.2\%$) compared to direct deployment of the off-the-shelf pre-trained models on ISP-less CV systems. Our approach, coupled with few-shot learning, has been shown to  be effective in detecting real raw objects captured directly by a camera from the PASCALRAW dataset.

\section{Acknowledgements}

We would like to acknowledge the DARPA HR$00112190120$ award and the NSF CCF-$1763747$ award for supporting this work. The views and conclusions contained herein are those of the authors and should not be interpreted as necessarily representing the official policies or endorsements of DARPA or NSF.

{\small
\bibliographystyle{ieee_fullname}
\bibliography{egbib}
}

\end{document}